\def\ARXIVVERSION{1}
\def\INCLUDESUPPLEMENT{1}
\documentclass[letterpaper]{article} 
\ifdefined\ARXIVVERSION
\usepackage[preprint]{aaai2027}
\else
\usepackage[submission]{aaai2027}  
\fi
\usepackage[hyphens]{url}  
\usepackage{graphicx} 
\usepackage{natbib}  
\usepackage{caption} 
\usepackage{algorithm}
\usepackage{algorithmic}

\usepackage{newfloat}
\usepackage{listings}
\DeclareCaptionStyle{ruled}{labelfont=normalfont,labelsep=colon,strut=off} 
\lstdefinestyle{promptbox}{%
	basicstyle={\footnotesize\ttfamily},
	numbers=none,xleftmargin=6pt,xrightmargin=6pt,
	backgroundcolor=\color{black!5},
	frame=single,framerule=0pt,framesep=6pt,
	columns=fullflexible,keepspaces=true,
	aboveskip=4pt,belowskip=2pt,
	showstringspaces=false,tabsize=2,breaklines=true}
\floatstyle{ruled}
\newfloat{listing}{tb}{lst}{}
\floatname{listing}{Listing}

\usepackage{booktabs}
\usepackage{array}      
\usepackage{amsmath}
\usepackage{amssymb}
\usepackage{placeins}

\usepackage{tikz}
\usetikzlibrary{arrows.meta,positioning}
\usepackage{pgfplots}
\pgfplotsset{compat=1.17}

\ifdefined\ARXIVVERSION
\newcommand{\extendedmaterial}{appendix}
\else
\newcommand{\extendedmaterial}{supplementary material}
\fi

{}\title{RMSWeb: Reflection, Failure-Mode Mining, and Salvage-DS for Web Agent Reinforcement Learning}
\ifdefined\ARXIVVERSION
\author{
Chengbo Liu\textsuperscript{\rm 1}\equalcontrib,
Lifang Zhou\textsuperscript{\rm 1}\equalcontrib,
Ruijie Yan\textsuperscript{\rm 1}\equalcontrib,
Pei Tan\textsuperscript{\rm 1},
Ao Sun\textsuperscript{\rm 2},\\
Haojun Huang\textsuperscript{\rm 1},
Guichun Hua\textsuperscript{\rm 1},
Sining Wei\textsuperscript{\rm 1},
Yining Chen\textsuperscript{\rm 1},
Yingying He\textsuperscript{\rm 1}\corresponding,
Yutao Xie\textsuperscript{\rm 1}\corresponding
}
\affiliations{
\textsuperscript{\rm 1}Microsoft, Beijing, China\\
\textsuperscript{\rm 2}Southeast University, Nanjing, China\\
chengboliu@microsoft.com, yingyhe@microsoft.com, yutaoxie@microsoft.com
}
\else
\author{Anonymous Submission}
\affiliations{}
\fi

\begin{document}

\ifdefined\SUPPLEMENTONLY
\maketitle
\section*{Supplementary Material}
\else
\maketitle

\begin{abstract}
Compact web agents can reduce deployment cost, but training them poses
challenges in both data collection and post-SFT reinforcement learning (RL).
Successful trajectories are expensive to collect and often contain inefficient
detours. After supervised fine-tuning (SFT), full trajectory corpora are
dominated by routine states; moreover, when group-relative RL is applied to web
actions, inadequately designed action-level rewards can yield weak or misleading
relative updates, while groups rejected as unsuitable for such updates receive
no fallback learning signal. We present
RMSWeb, a three-part recipe for Qwen3-VL-Instruct at 8B and 32B.
Reflection-conditioned retries increase collection yield and shorten successful
trajectories; failure-mode mining concentrates offline RL on critical states
exposed by the SFT policy; and Salvage-DS combines an
action-semantic polarized reward, contrast-and-competence-gated dynamic
sampling, and action-only anchor for rejected groups. Policies trained with
reflection-collected data use up to 19.7\% fewer action steps on solved tasks.
On WebVoyager, Online-Mind2Web, and WebTailBench, RMSWeb improves over
SFT by 2.4--7.0 points at 8B and 1.2--7.7 points at 32B. Our 8B model also
achieves the strongest reported Online-Mind2Web
result among similarly sized open-weight models in our comparison and a leading
reported accuracy--cost trade-off on WebVoyager and WebTailBench, with the
caveat that external evaluation protocols differ.
\end{abstract}


\section{Introduction}
\label{sec:introduction}

\IfFileExists{Figures/intro_accuracy_cost_tradeoff.pdf}{
\begin{figure*}[t]
\centering
\includegraphics[width=\textwidth]{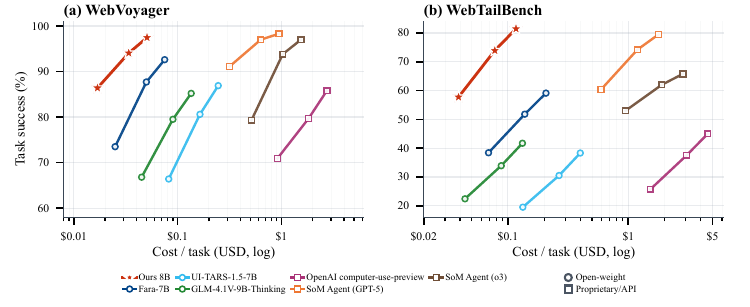}
\caption{Pass@$k$ accuracy--inference-cost trade-offs on WebVoyager and WebTailBench. Curves show independent-rollout pass@1/2/3. Our 8B model uses token-accounted single-run costs; external single-run accuracies and costs are reported by Fara-7B, with pass@2/3 accuracies reconstructed from its figures. All repeated-rollout costs use the nominal budget $C_k=kC_1$ without early stopping. Evaluation protocols differ across sources.}
\label{fig:intro-accuracy-cost}
\end{figure*}
}{}

Web agents use language or vision-language models to interpret browser
observations---screenshots, DOM/HTML, and accessibility trees---and emit
structured actions such as clicking or typing into page elements, scrolling,
switching tabs, and navigating to URLs
\citep{zhang2024llmbrained,nguyen2025guiagents}. Large proprietary
agents can be capable but expensive: repeated inference over long interaction
traces and multiple attempts raises cost and latency, motivating compact
open-weight alternatives.

Training compact agents exposes two broad challenges. First, successful
multi-step trajectories are expensive to collect and often contain avoidable
detours, so training data must improve both success yield and step efficiency
\citep{nakano2021webgpt,gur2024webagent,lu2024weblinx}. Second, improving a
strong SFT policy through offline RL requires effective data selection and
optimization. Most trajectory states may already be routine for the policy,
while supervised action-level rewards and group-relative updates can provide
weak or unreliable guidance on the remaining failures. We study these
challenges through a shared Browser Use
interface for collection, training-data generation, and live-web evaluation
\citep{browseruse2024}.

RMSWeb addresses the data challenge through reflection-conditioned collection
and failure-mode mining, and the optimization challenge through Salvage-DS.
Reflection improves collection yield and shortens successful trajectories;
mining selects critical states exposed by the SFT policy; and Salvage-DS admits
relative RL only for groups with reward contrast and a competent rollout while
routing rejected groups to action-only supervision.

We train offline by scoring sampled actions against verified targets. RMSWeb
improves over SFT on WebVoyager, Online-Mind2Web, and WebTailBench at both model
scales, while reflection-collected data reduces action steps under matched
decoding. Under our fixed live-web evaluation protocol, the 8B model achieves
86.39\% on WebVoyager and 57.74\% on WebTailBench; in the source-reported
accuracy--cost comparison of Figure~\ref{fig:intro-accuracy-cost}, these results
place it on the leading frontier.

Together, these results support three contributions:
\begin{enumerate}
	\item A reflective, multi-round data-collection pipeline that, for seed
	tasks, yields more successful trajectories and more
	step-efficient ones --- so the trained policy completes tasks in up to
	19.7\% fewer action steps under matched decoding.
	\item A failure-mode-guided miner that turns the SFT model's own failures into
	a small, targeted RL corpus of failure-critical steps (with
	ground-truth actions), instead of running RL over the full dataset.
	\item Salvage-DS, a step-level RL algorithm whose three components fix the ways
	group-relative RL stalls on a strong SFT policy: a polarized
	high-variance reward, a contrast-and-competence-gated dynamic sampler, and a
	salvage anchor that turns the groups the sampler
	rejects from RL into supervised teaching on hard states and anti-forgetting on solved
		ones, all in one objective.
\end{enumerate}

\section{Related Work}
\label{sec:related-work}

{}\textbf{Web and GUI agents.} Web agents specialize (M)LLM-driven GUI agents
for browser observation, grounding, and action
\citep{zhang2024llmbrained,nguyen2025guiagents}. Interfaces range from
HTML/DOM to screenshots with Set-of-Mark or learned visual grounding
\citep{yang2023setofmark,cheng2024seeclick,gou2025uground}; systems such as
SeeAct and UI-TARS integrate these capabilities end to end
\citep{zheng2024seeact,qin2025uitars}. Test-time methods instead improve a
fixed policy through reasoning, reflection, search, or workflow memory
\citep{yao2023react,shinn2023reflexion,koh2024treesearch,gu2024webdreamer,wang2024awm}.
We use reflection only to collect training data, update the weights, and retain
single-rollout deployment through Browser Use \citep{browseruse2024}.

{}\textbf{Training data: collection, synthesis, and curation.} Human
demonstrations are costly, motivating trajectory synthesis from indirect
knowledge or tutorials
\citep{ou2024synatra,xu2025agenttrek}, reverse task synthesis
\citep{sun2025osgenesis}, and environment exploration
\citep{murty2025nnetnav,pahuja2025explorer}. Related systems generate curricula,
tasks, or interaction data for self-improvement
\citep{qi2025webrl,bai2024digirl,putta2024agentq,he2024openwebvoyager,zhou2024pae,su2025learnbyinteract}.
These methods primarily expand data coverage or volume; RMSWeb instead uses
reflection to increase yield while shortening successful trajectories, then
mines the current SFT policy's failures into a compact critical-state curriculum.

{}\textbf{Learning web-agent policies.} Demonstration-based SFT underlies
WebGPT, WebAgent/HTML-T5, and WebLINX
\citep{nakano2021webgpt,gur2024webagent,lu2024weblinx}, although WebGPT also
uses rejection sampling against a human-preference reward model. Beyond
imitation, AutoWebGLM bootstraps with rejection sampling and RL
\citep{lai2024autowebglm}, and WebAgent-R1 applies end-to-end multi-turn RL
\citep{webagentr12025}; data-centric RL systems are summarized above. Our
offline objective is closest to step-level action-reward methods: GUI-R1 uses
verifiable rewards over GUI action components, while SRL provides dense
similarity-based rewards from expert step trajectories
\citep{luo2025guir1,deng2026srl}. We adopt action
matching rather than claim it as new; our contribution is competence-aware
routing and action-only salvage, building on PPO, GRPO, and DAPO
\citep{schulman2017ppo,shao2024deepseekmath,yu2025dapo}.

{}\textbf{Benchmarks and evaluation.} Evaluation spans offline action
prediction (Mind2Web), self-hosted execution (WebArena and VisualWebArena),
live-web tasks (WebVoyager and Online-Mind2Web), and full-OS control (OSWorld)
\citep{deng2023mind2web,zhou2024webarena,koh2024visualwebarena,he2024webvoyager,xue2025illusion,xie2024osworld}.
BrowserGym unifies many environments \citep{dechezelles2024browsergym}. We use
three live-web benchmarks and treat externally reported computer-use systems
only as protocol-qualified reference points \citep{awadallah2025fara}.

\section{Reflective Collection and Failure-Critical Data Mining}
\label{sec:data}

Naive trajectory collection has two shortcomings: independent teacher attempts
often fail, and even successful runs can contain avoidable detours.
We therefore condition each retry on a reflection distilled from the preceding
attempt. This turns prior experience into recovery guidance, increasing the
yield of usable successful trajectories while steering the teacher toward
shorter paths for SFT.

RL data pose a different problem: most trajectory steps are routine states
already mastered by SFT, so training on the full corpus dilutes the sparse
decisions where the policy still fails. We therefore contrast
same-task executions and mine critical steps where a successful trajectory and a
failed/inefficient trajectory diverge. The resulting compact curriculum uses successful
executions to specify what action to learn, while failed or inefficient
executions identify where RL training should be concentrated.

\subsection{Seed tasks and reflective trajectory collection}
\label{sec:seed-tasks}

Starting from a pool of seed tasks, this pipeline builds an SFT corpus covering 5K+
tasks through seed generation, reflection-conditioned collection, and quality
filtering. Relative to otherwise identical non-reflective collection,
reflection raises usable-success yield and shortens collected trajectories; the
reflective-collection yield analysis in the \extendedmaterial{} quantifies the yield, and the
reflection ablation (Table~\ref{tab:reflection-efficiency}) shows the shorter
trajectories carrying through to fewer action steps in the trained policy.

\begin{figure*}[t]
\centering
\includegraphics[width=\textwidth]{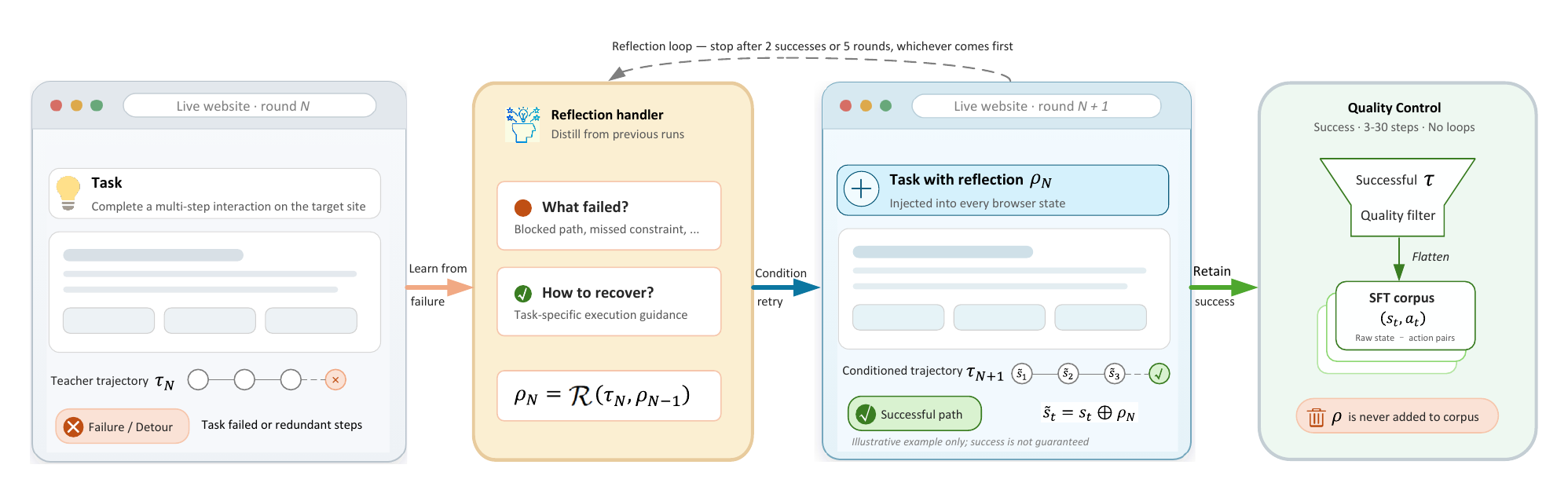}

\caption{Reflective trajectory collection. A reflection handler distills current
trajectory $\tau_N$ and the previous reflection into a task-specific guidance,
which is injected into every
browser state of the next round, but never added to the final corpus.}
\label{fig:reflection-loop}
\end{figure*}

{}\textbf{Seed-task generation.} We sample anonymized user search queries from
a commercial web search platform, without manual curation or difficulty
filtering, and use an LLM to transform them into executable web-agent tasks
with explicit websites, objectives, and constraints.

{}\textbf{Reflection-conditioned collection.} Our teacher model, GPT-5.1,
produces a
trajectory $\tau_N$; a reflection handler then summarizes its failures and
recovery strategy, together with the previous reflection, into $\rho_N$; the
full prompt appears in the \extendedmaterial. The
next round conditions every state on this reflection:
\[
\begin{array}{l@{\;}c@{\;}l}
\tau_N & = & (s^{N}_1,a^{N}_1,\ldots,s^{N}_{T_N},a^{N}_{T_N},\mathrm{done}),\\
\rho_N & = & \mathcal{R}(\tau_N,\rho_{N-1}),\\
\tilde{s}^{N+1}_t & = & s^{N+1}_t\oplus\rho_N.
\end{array}
\]
where $s_t^N$ and $a_t^N$ denote the observation and structured teacher action
at step $t$ of round $N$; $T_N$ is the trajectory length and
$\rho_0=\varnothing$. Here $\tilde{s}^{N+1}_t$ is the reflection-conditioned
observation and $\oplus$ denotes context concatenation.
Figure~\ref{fig:reflection-loop} illustrates this loop, which stops after two
successes or five rounds. Reflection guides only
collection: we discard $\rho_N$ when constructing SFT records and retain the
raw $(s^N_t,a^N_t)$ pairs from successful trajectories.

{}\textbf{Quality filtering.} We retain successful trajectories with complete,
valid step records, no stalled action loops, and 3--30 effective steps, thereby
excluding trivial or pathologically long runs.

{}\textbf{Corpus and flattening.}
The \extendedmaterial{} reports the retained trajectories'
distribution across eleven
top-level activity categories. Flattening the trajectories produces 150k+
per-step action and extraction samples for SFT.

\subsection{Failure-mode-guided critical-step mining}
\label{sec:critical-step-mining}

The full trajectory distribution is a poor RL curriculum because routine
transitions dominate the comparatively few failure-critical decisions.
We instead compare
executions of the same task to find the last state-aligned action divergence
whose surrounding evidence matches an assigned failure mode, then retain only
the successful-side decision. Figure~\ref{fig:rl-data-generation} illustrates
the critical step identification process.

\begin{figure}[t]
\centering
\includegraphics[width=\columnwidth]{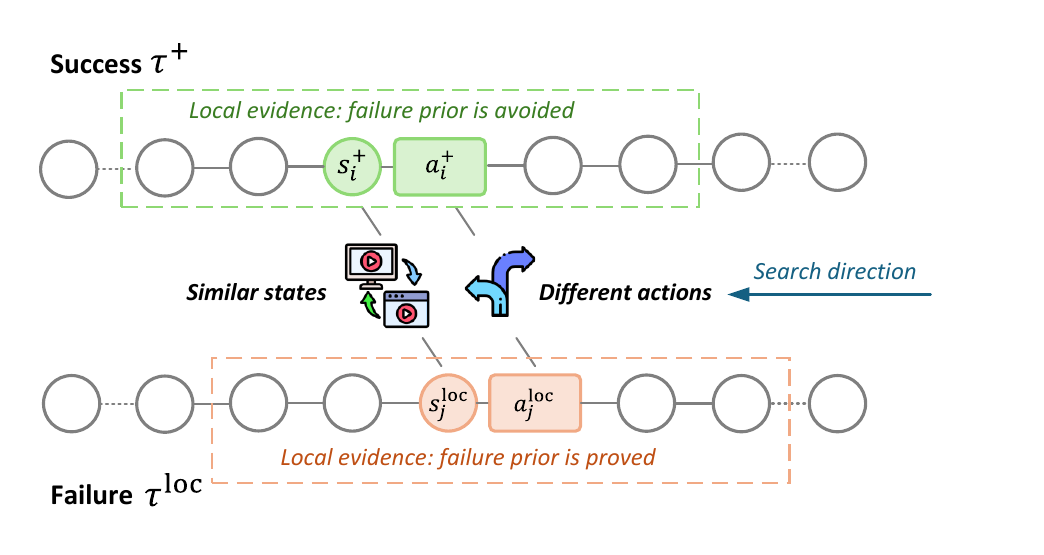}
\caption{Failure-mode-guided RL data generation. For the same task, a
verified successful trajectory $\tau^{+}$ is contrasted with a localization
trajectory $\tau^{\mathrm{loc}}$ that fails or takes a detour. We search the last
critical divergence, where the two steps have comparable states,
different actions, and local evidence that explains the failure.}
\label{fig:rl-data-generation}
\end{figure}

{}\textbf{Validation-derived failure prior.} We analyze the frozen SFT policy on
a held-out development split containing neither benchmark tasks nor
benchmark-specific action rules, yielding the task-agnostic taxonomy reported
in the \extendedmaterial. For each same-task contrast,
an LLM judge assigns a category and confidence from the task, actions,
and terminal summaries. This label acts only as a
soft search prior, favoring local windows where the successful branch avoids
the failure mode and the localization branch exhibits it.

{}\textbf{Divergence search.} The mining pool
augments the SFT collection with unsuccessful runs that SFT filtering discards.
We primarily pair a verified success $\tau^{+}$ with an unsuccessful execution
$\tau^{\mathrm{loc}}$ of the same task; as a secondary signal, we pair the
shortest success with longer successes to expose detours. We enumerate
non-terminal step pairs from late to early and require comparable pre-action
states---measured by URL/path, visible DOM elements, agent memory, and normalized
progress---but different ensuing actions. After removing search-only and
wait/scroll/back-only differences, we rank candidates by state alignment,
action contrast, and local failure-mode evidence, retaining the highest-scoring
pair that passes validity, visual-context, and progress gates. This alignment
supports trajectories with different lengths and navigation routes; the
\extendedmaterial{} gives the scoring details.

{}\textbf{Quality control.} We remove malformed,
low-scoring, trivial, or visually unusable pairs, balance categories and
domains, apply stricter gates to runtime/evaluation failures, and cap technical
obstructions at 10\%. This yields 4.5k critical-state examples. For the selected
divergence $(i^{*},j^{*})$, only $(s^{+}_{i^{*}},a^{+}_{i^{*}})$ enters RL:
GRPO samples actions conditioned on $s^{+}_{i^{*}}$ and scores them against the
verified $a^{+}_{i^{*}}$. The localization branch is only for data generation,
not used in the RL stage.

\section{The Salvage-DS RL Algorithm}
\label{sec:training}

We first SFT Qwen3-VL-8B-Instruct and Qwen3-VL-32B-Instruct on the successful
trajectories, then initialize RL from the best SFT checkpoints. Training
details are in the \extendedmaterial.

Training is offline: at each mined state, the policy samples actions and is
scored against a verified target without live-browser execution. Online
outcome-based RL is left to future work. Because web actions are structured,
exact matching can erase meaningful near-misses, while generic text similarity
can reward a wrong element, value, or URL. We therefore use a polarized
action-semantic reward tailored to web interaction; the \extendedmaterial{}
gives the action-specific rules. Salvage-DS combines this reward with
contrast-and-competence-gated dynamic sampling, which rejects uninformative or untrustworthy
groups from relative RL, and an action-only salvage anchor for retained
rejected groups. The complete objective below gives the joint
objective.
Figure~\ref{fig:salvage-ds} maps each component to the pathology it repairs;
Figure~\ref{fig:salvage-ds-algorithm} shows the complete training flow.

\begin{figure}[t]
\centering
\includegraphics[width=\columnwidth]{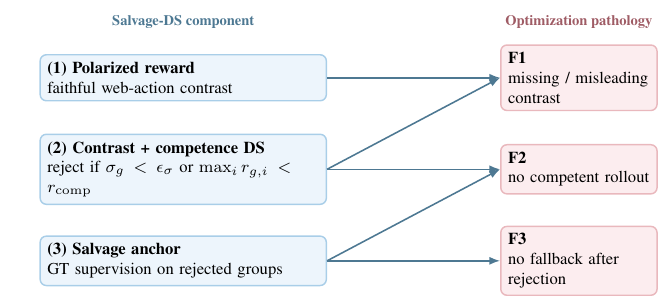}
\caption{Salvage-DS maps each optimization pathology to the component that
repairs it. \textbf{(1)} A web-action-semantic polarized reward restores
faithful contrast when generic rewards erase or misrank action differences
(F1). \textbf{(2)} Contrast-and-competence-gated dynamic sampling \emph{rejects} residual
low-contrast groups (F1) and groups with no
competent rollout (F2), rather than reinforcing an untrustworthy positive.
\textbf{(3)} The salvage anchor supplies a fallback gradient to retained
rejected groups---correcting unsolved states and anchoring mastered ones (F3).}
\label{fig:salvage-ds}
\end{figure}

\begin{figure*}[t]
\centering
\includegraphics[width=\textwidth]{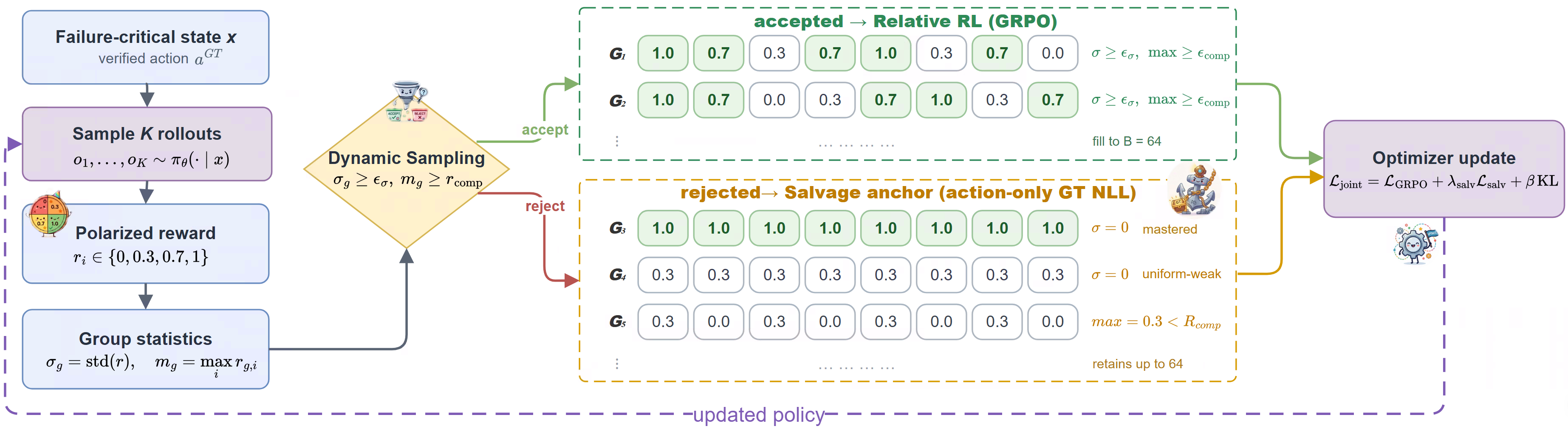}
\caption{\textbf{Salvage-DS training flow.} For each failure-critical state,
the policy samples $K$ actions, applies the polarized reward, and computes
group std $\sigma_g$ and maximum reward $m_g=\max_i r_{g,i}$. Groups passing
both gates are accepted into relative RL; rejection by either gate routes the
group to the action-only salvage anchor. Refill supplies exactly 64 accepted
groups, whereas the salvage branch retains up to 64 rejected groups, rounded
down to a multiple of 16. The relative-RL, salvage, and KL terms form the joint
update, after which the updated policy is sampled again.}
\label{fig:salvage-ds-algorithm}
\end{figure*}

\subsection{Group-relative optimization pathologies}
\label{sec:rl-motivation}

Our SFT model already achieves pass@3 $\gg$ pass@1, so a natural goal of RL is
to convert multi-sample competence into single-run reliability. For a group of
$K$ rollouts, GRPO uses the relative advantage
$\hat A_i=(r_i-\mu_g)/\sigma_g$, which exposes three optimization pathologies in web-agent training:

\textbf{(i) Unfaithful reward contrast gives missing or misleading feedback.}
Exact matching maps distinct near-misses to the same score, eliminating relative
signal; generic similarity can instead rank a semantically wrong web action too
high. Even with a faithful reward, uniform policy samples retain no contrast
\citep{yu2025dapo,deng2026srl}.

\textbf{(ii) No competent rollout gives no trustworthy positive.} If every
action is wrong but rewards differ, the highest-reward wrong action still receives positive
relative advantage.

\textbf{(iii) Rejected groups lack a fallback signal.} Without salvage,
unsolved groups receive no correction and mastered groups receive no retention,
so the former remain unlearned while the latter can drift under shared updates.

Salvage-DS therefore accepts relative RL only when a group has both reward
contrast and a competent rollout; rejected groups receive ground-truth action
supervision, correcting hard states and anchoring mastered ones.

\subsection{Polarized step-level reward}
\label{sec:polarized-reward}

Binary exact matching aliases near-misses with failures, whereas generic string
similarity can reward semantically wrong actions (e.g., adjacent click indices
or a nearly identical URL with the wrong date). We instead use a
polarized, action-semantic reward: action-specific rules map outputs to
1.0 (exact), 0.7 (core-correct), 0.3 (right action
type but wrong key parameter), or 0.0 (invalid/wrong action). This
preserves meaningful contrast without relying on surface similarity; the full
rubric and spacing analysis are in the \extendedmaterial.

\subsection{Contrast- and competence-gated Dynamic Sampling}
\label{sec:dynamic-sampling}

DS admits relative feedback only when it is both \emph{informative} and
\emph{trustworthy}. For rewards $\{r_{g,i}\}_{i=1}^{K}$, the std gate requires
within-group contrast ($\sigma_g\ge\epsilon_\sigma$), while the competence gate
requires at least one competent rollout ($\max_i r_{g,i}\ge r_{\mathrm{comp}}$):
\begin{equation}
\label{eq:ds-mask}
\begin{array}{l@{\;}c@{\;}l}
k_g & = & \underbrace{\mathbf{1}\!\left[\sigma_g\ge\epsilon_{\sigma}\right]}_{\text{std gate}}\cdot\underbrace{\mathbf{1}\!\left[\textstyle\max_i r_{g,i}\ge r_{\mathrm{comp}}\right]}_{\text{competence gate}},\\[2pt]
\hat A_{g,i} & = & k_g\,(r_{g,i}-\mu_g)/(\sigma_g+\varepsilon).
\end{array}
\end{equation}
Here $\sigma_g$ is the Bessel-corrected sample standard deviation and
$\varepsilon=10^{-6}$ stabilizes normalization. A group rejected by either
gate can have nonzero raw normalized advantages, but the mask sets every gated
$\hat A_{g,i}$ to zero; it contributes neither relative-RL loss nor gradient.
Refill generates until 64 groups are accepted; up to 64 rejected groups are retained for salvage
(see the configuration in the \extendedmaterial). Unlike binary-count filtering in DAPO
\citep{yu2025dapo}, the graded std gate detects uniform intermediate outcomes,
and the competence gate supplies an absolute correctness bar. Both use raw,
KL-free rewards.

\subsection{The salvage anchor on rejected groups}
\label{sec:salvage-ds}

For retained rejected samples $\mathcal{D}_{\mathrm{rej}}$, the salvage anchor
applies ground-truth NLL only over the action span $A_i$:

\[
\mathcal{L}^{\mathrm{salv}}(\theta)=-\,\mathbb{E}_{i\in\mathcal{D}_{\mathrm{rej}}}\!\Big[\tfrac{1}{|A_i|}\!\sum_{t\in A_i}\log \pi_\theta\big(a^{\mathrm{GT}}_{i,t}\mid \cdot\big)\Big].
\]

This term is weighted by $\lambda_{\mathrm{salvage}}$ in the joint objective.
Its cross-entropy gradient strongly teaches hard states whose target action is
unlikely, but is naturally small on confident mastered states, where it anchors
against forgetting. The full reasoning-plus-action response remains in context,
but only action tokens receive loss, correcting the decision without imposing
the teacher's reasoning style. The routing analysis in the \extendedmaterial{} gives the resulting
teach-while-doing dynamics and per-group gradient.

\subsection{Complete objective}
\label{sec:implementation-summary}

Collecting the polarized reward $R$, dynamic-sampling advantage $\hat A$, and
salvage anchor $\mathcal{L}^{\mathrm{salv}}$, the actor minimizes
\begin{equation}
\label{eq:objective}
\begin{aligned}
\mathcal{L}(\theta) =\ & -\,\mathbb{E}_{i\in\mathcal{A},\,t}\!\left[\min\!\big(\rho_{i,t}\hat A_i,\ \rho^{\mathrm{clip}}_{i,t}\hat A_i\big)\right]\\
& +\ \lambda_{\mathrm{salvage}}\,\mathcal{L}^{\mathrm{salv}}(\theta)\ +\ \beta\,\mathrm{KL}(\pi_\theta\Vert\pi_{\mathrm{ref}}),
\end{aligned}
\end{equation}
where $\rho_{i,t}=\pi_\theta(o_{i,t})/\pi_{\theta_{\mathrm{old}}}(o_{i,t})$ is the
importance ratio, $\rho^{\mathrm{clip}}_{i,t}=\mathrm{clip}(\rho_{i,t},1{-}\epsilon_{\mathrm{lo}},1{+}\epsilon_{\mathrm{hi}})$
its clipped form with asymmetric range
$[\epsilon_{\mathrm{lo}},\epsilon_{\mathrm{hi}}]=[0.2,0.32]$. The dynamic-sampling
acceptance mask $k_g$ (Eq.~\ref{eq:ds-mask}) partitions the sampled rollouts into the
\emph{accepted} set $\mathcal{A}=\{i:k_{g(i)}{=}1\}$ (informative groups, whose gated
advantage $\hat A_i$ drives the first term) and the \emph{rejected} set
$\mathcal{D}_{\mathrm{rej}}=\{i:k_{g(i)}{=}0\}$ (the same set salvaged by
$\mathcal{L}^{\mathrm{salv}}$). Here $\pi_{\mathrm{ref}}$ is the frozen SFT
policy. Thus each rollout contributes to either the GRPO/PPO-clip surrogate or
the salvage anchor, with a standard KL brake. The corresponding per-group
gradient is derived in the \extendedmaterial.

We apply this objective to the mined corpus at both model scales. Standard
stabilizers and all hyper-parameters are detailed in the \extendedmaterial.

\begin{table*}[!t]
\centering
{\small
\setlength{\tabcolsep}{4pt}
\begin{tabular}{lccccc}
\csname toprule\endcsname
Model / System & Size / Access & Tasks (WV / OM2W / WTB) & WV & OM2W & WTB \\
\midrule
\multicolumn{6}{l}{\textit{Proprietary / API}} \\
SoM Agent (o3) \citep{awadallah2025fara} & Closed & F595 / 300 / 609 & 79.3 & 55.4 & 52.7 \\
SoM Agent (GPT-5) \citep{awadallah2025fara} & Closed & F595 / 300 / 609 & 90.6 & 57.7 & 60.4 \\
OpenAI computer-use-preview \citep{awadallah2025fara} & Closed & F595 / 300 / 609 & 70.9 & 42.9 & 25.7 \\
Gemini CUA preview \citep{gupta2026molmoweb} & Closed & F595 / 300 / 609 & 88.6$^{\dagger}$ & 57.3$^{\ddagger}$ & 63.0$^{\dagger}$ \\
Claude Sonnet 4 \citep{yan2026m2} & Closed & 12 sites / 291 / --- & 80.5$^{\ddagger}$ & 67.7$^{\ddagger}$ & --- \\
\midrule
\multicolumn{6}{l}{\textit{Open-weight models}} \\
UI-TARS-1.5-7B \citep{awadallah2025fara} & 7B & F595 / 300 / 609 & 66.4 & 31.3 & 19.5 \\
GLM-4.1V-9B-Thinking \citep{awadallah2025fara} & 9B & F595 / 300 / 609 & 66.8 & 33.9 & 22.4 \\
Fara-7B \citep{awadallah2025fara} & 7B & F595 / 300 / 609 & 73.5 & 34.1 & 38.4 \\
MolmoWeb-8B \citep{gupta2026molmoweb} & 8B & F595 / 300 / 609 & 78.2$^{\dagger}$ & 35.3$^{\dagger}$ & 49.5$^{\dagger}$ \\
Weblica-8B \citep{kar2026weblica} & 8B & --- / 300 / NR & --- & 39.2 & 33.5 \\
GUI-Owl-1.5-8B-Thinking \citep{xu2026mobileagent35} & 8B & NR / 300 / --- & 78.1$^{\ddagger}$ & 48.6$^{\ddagger}$ & --- \\
Qwen3-VL-235B-A22B-Thinking \citep{yang2026openwebrl} & 235B MoE (22B active) & F595 / 300 / --- & 66.4$^{\ddagger}$ & 63.7$^{\ddagger}$ & --- \\
\midrule
\multicolumn{6}{l}{\textit{Ours (same evaluation harness)}} \\
Qwen3-VL-8B-Instruct (base) & 8B & 588 / 300 / 609 & 51.13 & 36.56 & 23.31 \\
\quad + SFT & 8B & 588 / 300 / 609 & 82.65 & 59.44 & 55.39 \\
\quad + SFT + RL & 8B & 588 / 300 / 609 & \textbf{86.39} & \textbf{66.44} & \textbf{57.74} \\
Qwen3-VL-32B-Instruct (base) & 32B & 588 / 300 / 609 & 75.68 & 53.67 & 39.57 \\
\quad + SFT & 32B & 588 / 300 / 609 & 86.45 & 66.11 & 51.62 \\
\quad + SFT + RL & 32B & 588 / 300 / 609 & \textbf{87.64} & \textbf{68.11} & \textbf{59.28} \\
\bottomrule
\end{tabular}
}
\caption{Task success (\%) on WebVoyager (WV), Online-Mind2Web (OM2W), and WebTailBench (WTB).
Our results are means over three independent Pass@1 evaluations under one
shared protocol; bold marks the best result within each of our model sizes.
External results follow source protocols. $\dagger$ denotes a
reported mean over 3--5 evaluations; $\ddagger$ denotes a reported Pass@1 result
for which repeated-run averaging is not reported. NR and dashes denote unreported task counts and
results, respectively.}
\label{tab:main-results}
\end{table*}

\section{Experiments}
\label{sec:experiments}

\subsection{Setup and evaluation protocol}
\label{sec:setup}

We evaluate on WebVoyager \citep{he2024webvoyager},
Online-Mind2Web \citep{xue2025illusion}, and WebTailBench
\citep{awadallah2025fara}, three complementary live-web benchmarks, and report
LLM-judged task success (\%), primarily pass@1; pass@$k$ denotes success in at
least one of $k$ attempts. No training task matches a benchmark task under our
lexical-overlap tests. Dataset definitions, contamination checks, aggregation,
and task-retention details are in the \extendedmaterial.

We compare, at both 8B and 32B, the untuned
Qwen3-VL-Instruct base, the SFT model, and our full
SFT + RL model (with the ablations reported below). Table~\ref{tab:main-results} also
includes representative proprietary, open-weight, and larger-scale systems
reported in their source papers. The full protocol for our runs --- the fixed Browser Use harness,
per-benchmark judges, live-site freshness and task exclusions, and the decoding
configuration --- is detailed in the \extendedmaterial.
WebVoyager task counts differ by source: F595 is Fara's LLM-verified subset of
the original 643 tasks, while our Browser Use-based retained set contains 588
tasks after excluding 55 outdated or unverifiable task IDs.
For WebTailBench, we use the same DOM-aware judge pipeline as for
WebVoyager, as also done by MolmoWeb \citep{gupta2026molmoweb}, rather than
Fara's native precomputed-rubric verifier. The latter reads success evidence
primarily from screenshots, whereas our policy acts on a combined
screenshot~$+$~DOM observation. Evaluation details are in the \extendedmaterial.

\subsection{Main results}
\label{sec:main-results}

Table~\ref{tab:main-results} reports task success rate on the three benchmarks. SFT already lifts the
8B base substantially, and RL on the failure-mode-mined corpus adds a
further consistent gain across all three benchmarks, giving our best 8B model.

Failure-mode-mined RL improves all three benchmarks at both scales. Over SFT,
it adds 3.7 / 7.0 / 2.4 points at 8B and 1.2 / 2.0 / 7.7 points at 32B on
WebVoyager / Online-Mind2Web / WebTailBench, indicating that the recipe scales
with model size.

On Online-Mind2Web, our 8B and 32B models reach 66.44 and 68.11 under our
evaluation harness. Source-reported external values use different harnesses;
for context, Claude~Sonnet~4 reports 67.7 and Qwen3-VL-235B reports 63.7.
On WebVoyager and
WebTailBench, GPT-5-/Gemini-based agents reach higher absolute
pass@1, but at much higher inference cost; Figure~\ref{fig:intro-accuracy-cost}
shows our 8B agent on the best accuracy--cost frontier on both, matching or
exceeding those systems per dollar and, with a few repeated attempts,
reaching their accuracy at a fraction of the cost. (The Fara-7B numbers are
quoted from its paper~\citep{awadallah2025fara} under a different evaluation
harness and serve only as an external reference point; see the
evaluation protocol in the \extendedmaterial.) We run the ablations in
the ablation subsection on the 8B model.

\subsection{Ablation studies}
\label{sec:experiment-ablations}

\paragraph{RL recipe.}
Table~\ref{tab:ablations} compares the full 8B recipe with three controls.
Full-corpus RL replaces the mined curriculum with the full SFT-style dataset;
anchor-free RL sets \texttt{salvage\_weight} to zero. DAPO uses the same mined
data, polarized reward, and KL brake, but removes the competence gate and
salvage anchor and uses DAPO-style Clip-Higher.

\begin{table*}[t]
\centering
\begin{tabular}{llccc}
\csname toprule\endcsname
Variant & Change vs. full recipe & WebVoyager & Online-Mind2Web & WebTailBench \\
\midrule
Ours (full recipe) & --- & \textbf{86.39} & \textbf{66.44} & \textbf{57.74} \\
Full-corpus RL & No failure-mode mining & 84.41 & 60.78 & 52.99 \\
Anchor-free RL & \texttt{salvage\_weight} = 0 & 84.69 & 60.56 & 56.71 \\
DAPO & No competence gate or salvage anchor & 84.92 & 62.67 & 54.74 \\
\bottomrule
\end{tabular}
\caption{Ablation of the RL recipe (task success rate, \%).}
\label{tab:ablations}
\end{table*}

\begin{table*}[!t]
\centering
\small
\begin{tabular}{l cc cc cc}
\csname toprule\endcsname
& \multicolumn{2}{c}{WebVoyager} & \multicolumn{2}{c}{Online-Mind2Web} & \multicolumn{2}{c}{WebTailBench} \\
\cmidrule(lr){2-3}\cmidrule(lr){4-5}\cmidrule(lr){6-7}
8B SFT training data & Acc & Avg. steps (succ.) & Acc & Avg. steps (succ.) & Acc & Avg. steps (succ.) \\
\midrule
With reflection & 82.65 & \textbf{9.57} & 59.44 & \textbf{15.96} & 55.39 & \textbf{14.83} \\
Without reflection & 83.28 & 10.99 & 60.78 & 19.42 & 50.14 & 18.47 \\
\midrule
Steps saved & \multicolumn{2}{c}{12.9\%} & \multicolumn{2}{c}{17.8\%} & \multicolumn{2}{c}{19.7\%} \\
\bottomrule
\end{tabular}
\caption{Reflection ablation for 8B SFT. The two models use the same training
recipe and decoding settings, differing only in whether reflection is used
during trajectory collection. Steps are averaged over successfully completed
tasks; steps saved is
$1-\mathrm{steps}_{\mathrm{with}}/\mathrm{steps}_{\mathrm{without}}$.}
\label{tab:reflection-efficiency}
\end{table*}

Full-corpus RL trails by 2.0 / 5.7 / 4.8 points on WebVoyager /
Online-Mind2Web / WebTailBench, showing that targeted mining matters beyond
additional RL. Removing only the salvage anchor costs
1.7 / 5.9 / 1.0 points. DAPO trails by
1.5 / 3.8 / 3.0 points despite sharing the mined data and reward,
supporting the combined benefit of competence-aware routing and salvage. The
dual corrective and retention roles of the anchor are detailed in the
Salvage-DS routing analysis in the \extendedmaterial.

\paragraph{Reflection-conditioned collection.}
Holding the 8B SFT recipe fixed, we vary only whether its trajectories were
collected with reflection. Under the same evaluation settings
(Table~\ref{tab:reflection-efficiency}), reflection leaves accuracy essentially
unchanged on WebVoyager and Online-Mind2Web and improves WebTailBench by
$+5.3$ points, while reducing success-task steps by
12.9--19.7\% across all three benchmarks. This mirrors the shorter training
trajectories reflection collects (13.7 vs.\ 16.2 steps). Reflection's separate
effect on collection yield is reported in the reflective-collection yield
analysis in the \extendedmaterial. We therefore adopt reflection as our
default: it buys step efficiency and a clear WebTailBench gain at no accuracy
cost on the shorter-horizon WebVoyager and Online-Mind2Web.

\subsection{Training behavior}
\label{sec:training-behavior}

Logged reward-gate, action log-probability, KL, clipping, and rollout-perplexity
metrics reveal consistent behavior at both scales. The fixed mined curriculum
progressively moves from no-competence rejection toward mastered-state
rejection, while the policy reverses its initial preference for plausible but
wrong actions by raising the correct action's likelihood. KL, clipping, and
rollout perplexity remain bounded, indicating learning rather than unstable
self-sharpening. Full training-dynamics analysis is reported in the
\extendedmaterial.

\FloatBarrier

\section{Conclusion}
\label{sec:conclusion}

We presented RMSWeb, a three-part recipe --- reflective multi-round data
collection, failure-mode mining after SFT, and Salvage-DS offline RL --- for
training vision-language web agents at 8B and 32B. In our post-SFT offline RL
setting, the results show complementary benefits from selecting
failure-critical states and retaining supervision when a sampled group is
unsuitable for a relative update. Failure-mode mining identifies the critical
states, while Salvage-DS combines an action-semantic polarized reward,
competence-aware dynamic sampling, action-only anchoring, and a KL brake.
Across the three benchmarks, RMSWeb improves over SFT by 2.4--7.0 points at 8B
and 1.2--7.7 points at 32B. Ablations
show that the full method outperforms RL on the full corpus, anchor-free RL,
and a DAPO baseline using the same mined data and polarized reward.

We will release the trained model weights and inference framework. The training
data are undergoing compliance review and will also be released if approved.

{}\textbf{Limitations and future work.} Our RL is offline: rewards come from
step-level matching against ground-truth actions, not from executing the agent
in a live environment. This is simpler and cheaper but caps the ceiling --- the
reward is a proxy for task success rather than success itself, and it cannot
credit novel-but-valid strategies that diverge from the ground truth. The most
promising next step is online RL in a real browser with outcome-based
rewards, where the training signal is the evaluation metric; the cost is a
live, resettable environment with its latency, non-determinism, and safety
concerns. Other open items: the failure-mode
taxonomy and local evidence rules are heuristic and LLM-assisted, and learned (process) reward models or
trajectory-level objectives could replace hand-specified step matching.

\bibliography{aaai2027}
\fi

\ifdefined\INCLUDESUPPLEMENT
\appendix

\section{Evaluation protocol details}
\label{app:eval-protocol}

{}\textbf{Benchmarks.} WebVoyager contains 643 open-ended tasks across 15 live
websites and uses an LLM outcome judge \citep{he2024webvoyager}; our retained
live-task set is described below. Online-Mind2Web contains 300 tasks over 136
sites and uses the three-stage WebJudge protocol \citep{xue2025illusion}.
WebTailBench contains 609 complex tasks spanning 11 types, including
transactional, multi-step, compositional, and cross-site workflows
\citep{awadallah2025fara}.

{}\textbf{Evaluation harness.} We fix a
single evaluation harness: an instrumented fork of the open-source
Browser Use framework \citep{browseruse2024}. All of our evaluations run in a
live browser, with the fork driving a real desktop Chrome instance under 16-way
concurrency --- not an offline replay or sandbox. Every task is run exactly
three times.

{}\textbf{Metric.} We micro-average judge-determined task success over retained
tasks. A run succeeds only when the agent declares completion and passes the
benchmark-specific check. Pass@$k$ is success in at least one of $k$
independent attempts; main tables report pass@1. WebVoyager uses the retained
588-task set described below.

{}\textbf{Contamination check.} We compare every training task with all tasks
in WebVoyager (643), Online-Mind2Web (300), and WebTailBench (609) using exact
match, whole-question substring containment, and word-level 3-gram
Jaccard/containment. Thresholds are 0.4/0.6 and are tightened by $0.6\times$ for
same-domain URLs. No training task passes any threshold.

{}\textbf{Live-site freshness and exclusions.} WebVoyager and WebTailBench both
contain time-sensitive tasks (e.g., flight and hotel bookings tied to specific
dates); for both, we refresh the affected prompts to future dates so that the
tasks remain executable on the live sites, applying the identical date-refresh
procedure to each. For WebVoyager, our Browser Use-based retained set contains
588 tasks after excluding 55 of the original 643 task IDs as outdated and
unrepairable by date edits; we apply this selection consistently and report
scores on all 588 tasks. Within these 588,
the 43 Cambridge
Dictionary tasks are run as a separate pass---without the custom desktop-Chrome
user agent, whose presence trips their anti-bot defenses---and merged into the
final aggregation. WebTailBench's task
set is used in full (only its date-bound prompts are refreshed, as above), and
Online-Mind2Web is used without modification. This reflects an inherent
limitation of live-web evaluation: absolute scores can drift as sites change, so
all models we run are evaluated within the same time window.

{}\textbf{Judges.} We adapt each benchmark's evaluation flow to a common
GPT-4o judge served through a load-balanced API deployment.
Our WebVoyager evaluation follows the browser-use WebVoyager
judge~\citep{browseruse2024}: a multimodal outcome judge over the final
response and the last four screenshots.
Online-Mind2Web is scored with the official OSU-NLP WebJudge implementation
unchanged---its three-stage protocol (key-point identification, screenshot
relevance scoring, and final verdict) over up to 30 screenshots---with only the
judge model swapped to our common GPT-4o. WebTailBench uses the same WebVoyager
judge pipeline described above, as also done by MolmoWeb
\citep{gupta2026molmoweb}. We use this shared outcome judge instead of Fara's
native precomputed-rubric verifier, which infers success primarily from
screenshots. We justify this choice in the cross-harness check below.

\section{Cross-harness check under Fara's native judge}
\label{app:fara-native-readjudication}

WebTailBench originates from Fara's evaluation framework
\citep{awadallah2025fara}, whose native verifier reads success evidence
primarily from screenshots and retains only the top-$K$ of them. Our policy
acts on a combined vision~$+$~DOM observation, so the judge's input can be
misaligned with what our policy is optimized against: success evidence
our agent reads from the DOM, or from screenshots outside the top-$K$ window, can
be missed by the native verifier.

To characterize this misalignment, we re-adjudicated every trajectory
the native judge marked as a failure and identified the \emph{false negatives}
among them. These false negatives fall into three categories, whose shares we
report from the full re-adjudication:
\begin{itemize}
\item \textbf{DOM-only evidence ($\approx$42\%).} The screenshots alone are
insufficient to confirm success, but the agent's DOM observation
(\texttt{browser\_state}) contains the key facts that support its final
answer.
\item \textbf{Incomplete screenshot reading ($\approx$37\%).} The full screenshot
sequence already suffices to prove success, but the judge overlooks some of the
screenshots or later page states (e.g., evidence beyond the retained top-$K$
window).
\item \textbf{Judge reasoning error ($\approx$21\%).} The judge fabricates task
requirements, misinterprets available evidence, or misapplies its
unavailability / technical-block / critical-point policies---e.g., a genuinely
valid booking whose search legitimately returns no matches (which Fara's own
rubric scores as success) is marked as a failure.
\end{itemize}

As all three categories inflate the failure count independently of agent
behavior, we score WebTailBench with the WebVoyager-style DOM-aware judge in our
common harness rather than Fara's native verifier, so that our vision~$+$~DOM
policy is credited for DOM-surfaced success evidence.

\section{Reflective-collection yield}
\label{app:reflection-yield}

Figure~\ref{fig:reflection-yield} compares collection with and without
reflection on the round-1 failure tasks (1.5k+). Each round is
one teacher-guided retry, and a task stops after two successes or five rounds;
since all these tasks fail at round~1, the curves span rounds~2--5. We track
\emph{solve@2}, the cumulative fraction of tasks that have reached two successful
trajectories by each round---a task-level measure of how much usable training
data accrues and how fast. Reflection leads at every round, and its solve@2
advantage widens monotonically with each retry, reaching $+12.7$ points by
round~5 ($33.6$ vs.\ $20.9$). This yield advantage is a collection-side measure;
the trained-policy step-efficiency effect of reflection is reported separately in
the reflection ablation in the main paper.

\begin{figure}[t]
\centering
\includegraphics[width=\columnwidth]{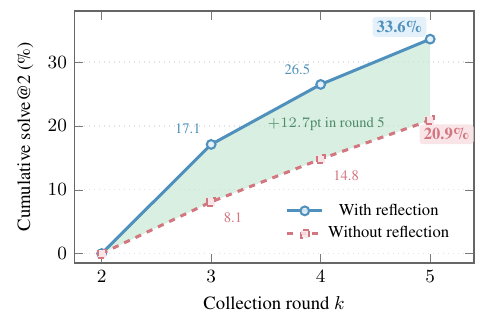}
\caption{Reflective-collection yield on round-1 failure tasks. \emph{solve@2}
is the cumulative fraction of tasks reaching two successful trajectories by
round~$k$; collection stops for each task after two successes or five rounds.
Blue $=$ with reflection, coral $=$ without.}
\label{fig:reflection-yield}
\end{figure}

\IfFileExists{Figures/Reflection_prompt.pdf}{
\section{Reflection prompt}
\label{app:reflection-prompt}

Figure~\ref{fig:reflection-prompt} shows the prompt used by the reflection
handler $\mathcal{R}(\tau_N,\rho_{N-1})$ of the reflective-collection subsection.
Given the current-round trajectory $\tau_N$ (rendered as a per-step summary of
URL, thinking, goal, and executed-action count) and, when available, the
previous-round guide $\rho_{N-1}$, it distills a compact guidance
$\rho_N$ that is injected into every browser state of the next round
and discarded when constructing SFT records.

\begin{figure*}[t]
\centering
\includegraphics[width=\textwidth]{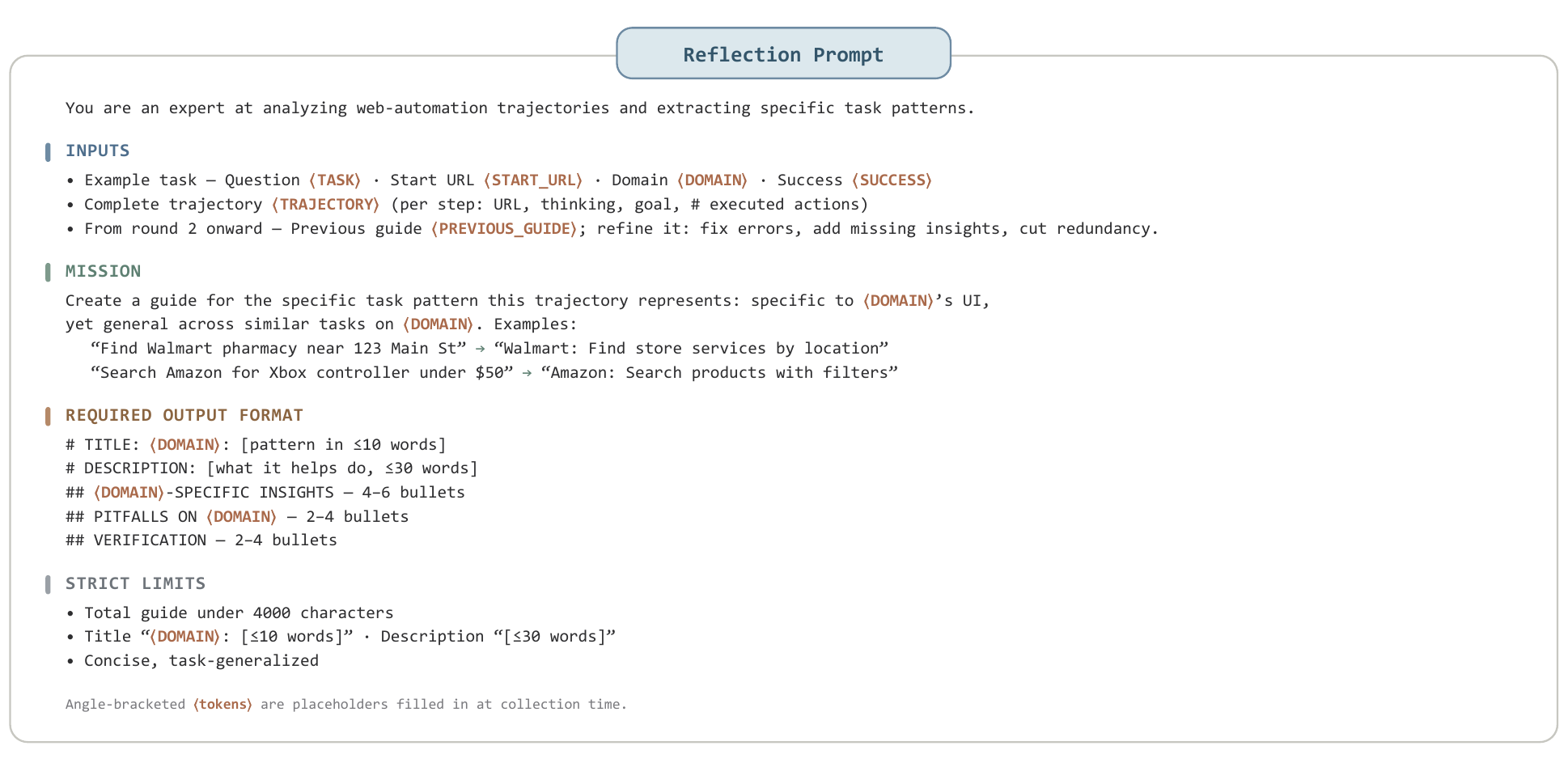}
\caption{Reflection prompt used during reflection-conditioned trajectory
collection. It turns one round's trajectory (and, from round two onward, the
previous round's guide) into a compact guidance $\rho_N$ that
conditions the next round; angle-bracketed tokens are placeholders filled in at
collection time.}
\label{fig:reflection-prompt}
\end{figure*}
}{}

\section{Task-query set distribution}
\label{app:task-distribution}

Figure~\ref{fig:task-distribution} summarizes the domain composition of the
task-query set used for data collection, aggregated into eleven top-level
categories. The distribution is heavy-headed and long-tailed: transactional
domains dominate, with E-Commerce (32.7\%), Travel \& Hospitality (21.0\%), and
Government (15.1\%) together accounting for roughly two-thirds of all tasks,
reflecting the everyday shopping, booking, and public-service intents that
anchor our anonymized search-log seeds. The remaining mass is spread across Entertainment \&
Gaming (5.3\%), Health (4.4\%), Corporate \& Careers (4.3\%), Finance (4.0\%),
Education (3.7\%), Real Estate (3.3\%), and Local \& Home Services (2.4\%), with
a small Other residual (3.9\%) grouping underrepresented micro-domains. This
spread keeps the corpus centered on high-frequency web activities while still
exercising a broad range of site types and interaction patterns.

\begin{figure*}[t]
\centering
\includegraphics[width=0.85\textwidth]{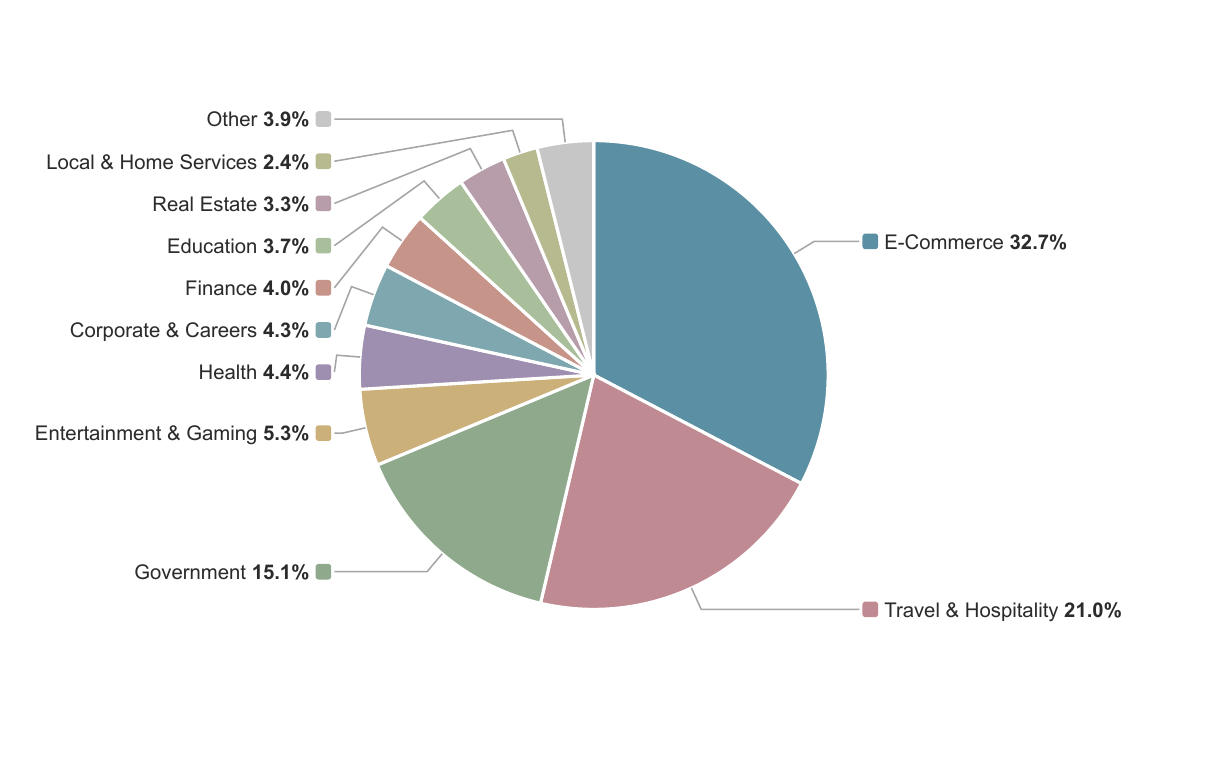}
\caption{Distribution of the task-query set by categories.}
\label{fig:task-distribution}
\end{figure*}

\section{SFT hyper-parameters}
\label{app:sft-hyperparameters}

\paragraph{Shared setup.}
We fine-tune \textbf{Qwen3-VL-8B-Instruct} and \textbf{Qwen3-VL-32B-Instruct} with LoRA (rank~64, $\alpha=128$,
dropout~0.05, targets = all linear layers), the \texttt{qwen3\_vl\_nothink} chat
template, a context window of 12{,}000 tokens, and an image cap of
262{,}144\,px. Optimization uses AdamW at learning rate $1\times10^{-4}$ with a
cosine schedule and weight decay~0.01, in bf16 for 3 epochs. Per-device batch
size is~1 and the random seed is~42. Both runs train on the reflective-collection dataset.

\paragraph{8B run.}
1~node with 8~GPUs. Gradient accumulation is~4 (effective batch~32) and warmup
ratio is~0.03.

\paragraph{32B run.}
2~nodes with 8~GPUs each (16~GPUs total). Gradient accumulation is~2
(effective batch~32) and warmup ratio is~0.1.

\section{Salvage-DS RL configuration}
\label{app:rl-hyperparameters}

\paragraph{Backbone and initialization.}
We tune LoRA of rank~64 and $\alpha=128$ (dropout~0.0) on targets
\texttt{linear\_qkv} and \texttt{linear\_proj}, initialized from the SFT model
with LoRA merged into the base weights.

\paragraph{Objective and reward.}
Group-relative advantages with dynamic sampling (\texttt{grpo\_ds}, GRPO + DS).
The reward uses the levels in the reward section of the \extendedmaterial{} via a custom
\texttt{compute\_score}. Dynamic sampling uses near-tie tolerance
$\epsilon_{\sigma}=0.10$ and minimum competent reward
$r_{\mathrm{comp}}=0.55$; the salvage weight is~0.10. The refill loop targets
64 informative groups with at most 16 generation batches per step.

\paragraph{Stability.}
Asymmetric PPO clipping $[0.2,\ 0.32]$; KL brake (\texttt{use\_kl\_loss},
coefficient $10^{-3}$, \texttt{low\_var\_kl}); no entropy bonus. Learning rate
$8\times10^{-6}$ with \texttt{seq-mean-token-mean} loss aggregation. These
choices suppress the self-sharpening pattern observed without the brake: the
policy increases the positive/negative log-probability gap by suppressing
negatives while KL, clipping, and rollout log-perplexity rise despite a reward
plateau. No KL term is included in the reward.

\paragraph{Batching and lengths.}
Train batch~64, mini-batch~16, dynamic batching capped at 20{,}480 tokens per
GPU. Rollouts per prompt $n=8$. Sequence lengths: prompt $\leq 12{,}000$,
response $\leq 2{,}048$. Refill always supplies 64 accepted groups. If $D$ rejected
groups are available, the salvage branch retains
$S=16\lfloor\min(D,64)/16\rfloor$ groups; hence $S\in\{0,16,32,48,64\}$ and
contains $8S\leq512$ rollout samples.

\paragraph{Data.}
The failure-critical mined corpus contains 4.5k
critical-step prompts.

\section{Salvage-DS reward details}
\label{app:reward}

Table~\ref{tab:reward-rules} jointly defines the four reward levels and their
representative web-action semantics. Exact matching alone would score a
formatting-only text variation like a wholly wrong input, erasing useful
contrast. Generic sequence similarity has the opposite failure: click indices
18 and 19 are lexically adjacent but can target different elements, while URLs
with the wrong date or values such as ``100'' and ``1000'' can remain highly
similar. We therefore first align action types, then compare task-critical
fields (element, content, destination, or success flag); text similarity is
used only within an already aligned semantic field. Multi-action steps aggregate
the resulting levels with weighted soft-F1.

\begin{table*}[t]
\centering
\small
\setlength{\tabcolsep}{3pt}
\begin{tabular}{>{\raggedright\arraybackslash}p{0.20\textwidth}>{\raggedright\arraybackslash}p{0.15\textwidth}>{\raggedright\arraybackslash}p{0.23\textwidth}>{\raggedright\arraybackslash}p{0.18\textwidth}>{\raggedright\arraybackslash}p{0.13\textwidth}}
\csname toprule\endcsname
Action & \textbf{1.0}: exact & \textbf{0.7}: core-correct & \textbf{0.3}: right type, wrong key parameter & \textbf{0.0}: wrong / invalid \\
\midrule
{}\shortstack[l]{\texttt{click\_element\_}\\\texttt{by\_index}} & all parameters match & index matches; secondary flag differs & index differs & wrong action / invalid \\
{}\texttt{scroll} & all parameters match & direction and frame match; magnitude differs & direction or frame differs & wrong action / invalid \\
{}\texttt{input\_text} & all parameters match & index matches; text similarity high & wrong index or dissimilar text & wrong action / invalid \\
{}\texttt{go\_to\_url} & all parameters match & host and path match; query may differ & same action, wrong destination & wrong action / invalid \\
{}\texttt{search\_engine} & query exactly matches & query similarity high & query similarity low & wrong action / invalid \\
{}\texttt{switch\_tab} / \texttt{close\_tab} & tab id matches & --- & tab id differs & wrong action / invalid \\
{}\texttt{done} & all parameters match & success flag matches; secondary field differs & success flag missing & success flag mismatches / invalid \\
\bottomrule
\end{tabular}
\caption{Polarized reward levels and representative per-action matching rules.
For every action, an exact parameter match receives 1.0; 0.7 preserves
semantically correct near-misses, 0.3 marks the right action type with a wrong
key parameter, and 0.0 denotes a wrong action or invalid output. Rules are
illustrative; the full implementation covers all action types.}
\label{tab:reward-rules}
\end{table*}

The polarized levels are spaced by at least 0.3. For a single action and eight
rollouts, a 7-to-1 split between adjacent
levels has Bessel-corrected sample std $0.3/\sqrt{8}\approx0.106$; an 8-to-0 split still
has zero std. Thus the spacing exposes semantic disagreements when present but
does not manufacture them. Uniform groups are handled by DS and the salvage
anchor.

\section{Salvage-DS routing}
\label{app:salvage-routing}

\paragraph{A teach-while-doing loop.} RL repeatedly samples the same mined
states, allowing the anchor and policy gradient to compound. When no rollout is
competent, the action-only anchor raises the likelihood of the ground-truth
action; the state can subsequently produce a competent rollout and graduate to
GRPO. On mastered states, the same cross-entropy gradient is small but anchors
the shared policy against forgetting.

\paragraph{Adaptive SFT--RL routing.} The mask $k_g$ sends informative groups
with a competent rollout to GRPO and all other groups to the action-only anchor.
Ignoring clipping notation, define
\begin{equation}
\label{eq:routed-gradient}
\begin{alignedat}{2}
&G_g &&= k_gG_g^{\mathrm{RL}}
 +(1-k_g)\lambda_{\mathrm{salvage}}G_g^{\mathrm{SFT}}
 +\beta G^{\mathrm{KL}},\\[2pt]
&G_g^{\mathrm{RL}} &&= -\mathbb{E}_{i,t\in g}
 [\hat A_{g,i}\nabla_\theta\log\pi_\theta(o_{i,t})],\\[2pt]
&G_g^{\mathrm{SFT}} &&= -\mathbb{E}_{i,t\in A_g}
 [\nabla_\theta\log\pi_\theta(a^{\mathrm{GT}}_{i,t}\mid\cdot)].
\end{alignedat}
\end{equation}
Hence $k_g=1$ gives relative RL, whereas $k_g=0$ supplies supervised
correction. The latter is strong when the ground-truth action is unlikely and
naturally weak on confident, mastered states. Thus, Salvage-DS dynamically
routes groups between relative RL and action-only supervision during training.

\section{Salvage-DS RL training dynamics}
\label{sec:training-dynamics}

\begin{figure*}[t]
\centering
\includegraphics[width=\textwidth]{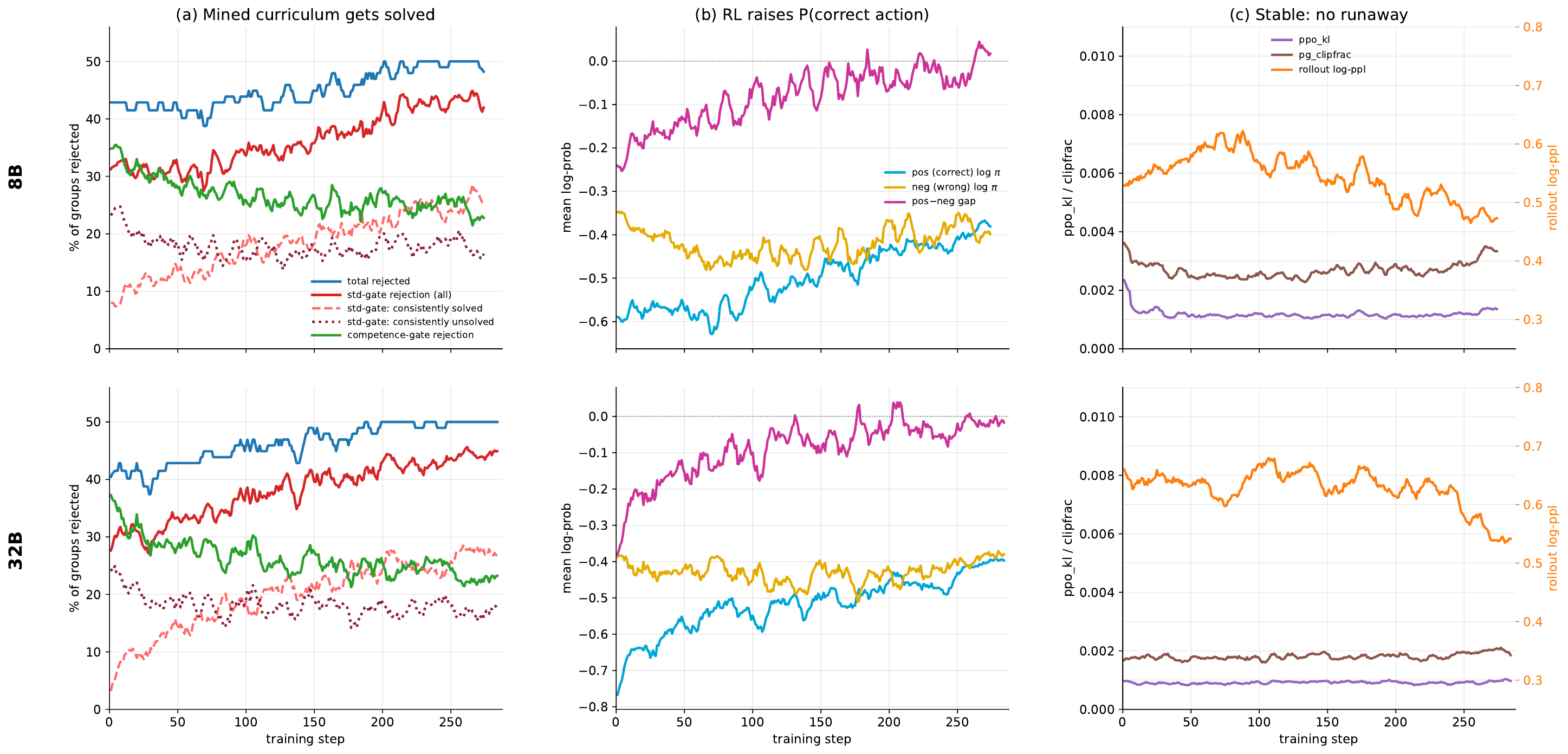}
\caption{Training dynamics at both scales (\textbf{top:} 8B; \textbf{bottom:}
32B). \textbf{(a)} The std-gate rejection share rises as mined states become
mastered, while competence-gate rejection falls. Red shades split std-gate
rejections into consistently solved ($\max_i r_{g,i}\ge r_{\mathrm{comp}}$)
and consistently unsolved ($\max_i r_{g,i}<r_{\mathrm{comp}}$) groups.
\textbf{(b)} Training raises the correct action's log-probability and reverses the initial
correct-minus-wrong gap. \textbf{(c)} KL, clipping, and rollout
log-perplexity remain bounded under the stabilized objective.}
\label{fig:dynamics}
\end{figure*}

\paragraph{Curriculum progression.} On the fixed mined states, the std-gate
rejection share rises from $\sim32\%$ to $\sim44\%$ at 8B, while
competence-gate rejection falls from $\sim31\%$ to $\sim21\%$. Concurrently,
best-of-8 reward increases ($0.63\!\rightarrow\!0.72$) and within-group std
decreases ($0.19\!\rightarrow\!0.14$). Together these trends indicate that
hard states progressively become competent and then mastered; refill maintains
64 accepted groups per update, while rejected groups are routed to salvage.

\paragraph{Targeted preference repair.} At the SFT initialization, the mined
states exhibit a negative correct-minus-wrong log-probability gap
($\approx-0.23$). Training raises the correct branch
($-0.59\!\rightarrow\!-0.37$), while the wrong branch changes little
($-0.36\!\rightarrow\!-0.41$), bringing the gap to $\approx0.04$. Thus the
policy repairs the targeted mis-preference rather than improving solely by
suppressing negatives.

\paragraph{Self-sharpening and stability.} Without the KL brake, the policy can
increase separation between sampled positive and negative branches much faster
than reward improves: the log-probability gap and
\texttt{rollout\_log\_ppl} grow, while reward improves only gradually and may
even temporarily regress. Concurrent sharp rises in \texttt{ppo\_kl} and
\texttt{pg\_clipfrac} indicate rapid policy drift and widespread clipping.
This joint pattern is the self-sharpening fingerprint: confidence and
distribution shift outpace task-reward progress. With the KL brake and zero
entropy bonus, Figure~\ref{fig:dynamics}c instead shows bounded KL, clip
fraction, and rollout log-perplexity.

\section{Critical-step mining details}
\label{app:critical-step-mining}

{}\textbf{Failure-label taxonomy.} The validation-derived taxonomy contains the
eleven operational categories in Table~\ref{tab:critical-step-failure-labels}.

\begin{table*}[t]
\centering
\small
\setlength{\tabcolsep}{3pt}
\begin{tabular}{>{\raggedright\arraybackslash}p{0.34\textwidth}r>{\raggedright\arraybackslash}p{0.51\textwidth}}
\csname toprule\endcsname
Category & Share & Explanation \\
\midrule
Unverifiable evidence, source, or compliance failure & 39.80\% & The conclusion lacks page evidence or relies on a source that does not satisfy the task. \\
Critical action or task endpoint incomplete & 18.73\% & The agent finds the target but does not add, book, authenticate, or reach the required confirmation state. \\
Access, anti-bot, or technical obstruction & 9.98\% & The trajectory stalls behind a CAPTCHA, 403 page, or unavailable control. \\
UI interaction or navigation failure & 8.53\% & A required detail page, tab, date picker, or dropdown is not reached or operated. \\
Runtime or evaluation failure & 7.38\% & The run times out, crashes, or loses an interpretable action trace. \\
Hard constraint, target object, or numeric threshold mismatch & 6.02\% & A selected option violates a required attribute or numeric threshold. \\
Filter/sort/search context not applied & 4.53\% & The trajectory claims a control is active, but the page state does not reflect it. \\
Date/time/availability closure & 2.26\% & A requested date, time, inventory, or recency condition is never verified. \\
Target discovery or search-strategy failure & 1.82\% & Repeated search or navigation choices do not reach the target page. \\
Information extraction, final calculation, or multi-field omission & 0.64\% & A required field, unit, comparison, or derived quantity is omitted. \\
Login, payment, or privacy-boundary error & 0.31\% & The agent stops at the wrong boundary or supplies unsupported sensitive information. \\
\bottomrule
\end{tabular}
\caption{Failure-mode distribution of the final corpus.}
\label{tab:critical-step-failure-labels}
\end{table*}

The labels identify where task closure breaks rather than prescribing
benchmark-specific behavior. Rule features provide an initial prior from task constraints, action types,
numeric and temporal cues, terminal behavior, and blocker signatures. An LLM
judge receives the task, summaries of the two trajectories' final steps, and
these features, then assigns a category and confidence. A second pass reviews
labels below confidence $0.70$; when the LLM confidence is very low, the
rule-derived category is used instead. In the final corpus, 4{,}356 labels
(95.74\%) come from the primary judgment and 194 (4.26\%) from this second-pass
review.

{}\textbf{Overall candidate score.} Every eligible state pair is ranked by the
same aggregate score used throughout mining:
\[
S_{\mathrm{total}}=0.50S_{\mathrm{state}}
+0.25S_{\mathrm{action}}+0.25S_{\mathrm{category}}.
\]
Here $S_{\mathrm{state}}$ measures whether the two branches present comparable
browser states, $S_{\mathrm{action}}$ measures whether their next actions form a
meaningful decision contrast, and $S_{\mathrm{category}}$ measures whether the
local divergence is supported by the assigned failure label. The following
paragraphs describe these components in implementation order.

{}\textbf{State alignment.} For each pair of non-terminal pre-action states,
the state component compares URL identity and path, overlap between visible DOM
elements represented as tag--text tuples, similarity of the agent memories, and
normalized progress through the two trajectories. We keep only pairs with a
state score of at least $0.60$, different action compositions, and at least one
non-trivial action. Pairs whose combined actions contain only search operations,
or only \texttt{wait}, \texttt{scroll}, and \texttt{go\_back}, are removed.

{}\textbf{Action contrast.} The action component increases when the branches
choose different action types or materially different parameters, and when at
least one branch performs a state-changing operation such as clicking, typing,
selecting, navigating, extracting, or completing. It decreases for
low-information differences dominated by passive navigation or repeated search.
This favors divergences that represent consequential decisions rather than
minor syntactic variation.

{}\textbf{Failure-mode evidence.} For the assigned failure category, we inspect
a radius-two window around each candidate on both branches. The category
component increases when the successful window exhibits the category's desired
closure behavior and the localization window exhibits the corresponding
violation. It decreases when the localization branch already contains the same
positive evidence, since that makes the contrast less diagnostic. The
trajectory-level label confidence serves as a soft prior, so local behavioral
evidence remains necessary even for a confident label.

{}\textbf{Selection and post-filtering.} We retain the highest-ranked candidate
with $S_{\mathrm{total}}\geq0.62$ and $S_{\mathrm{category}}\geq0.45$, valid
model input and output, usable visual context on at least one branch, and a
non-trivial successful-side progress action. Category-aware sampling and a 30\%
per-domain cap reduce source concentration. Runtime/evaluation examples must
satisfy stricter confidence and diagnostic-signal gates, while anti-bot and
technical-obstruction examples are downsampled to at most 10\% of the final
corpus.


{}\textbf{Qualitative positive/negative contrasts.} Table~\ref{tab:critical-step-cases}
shows retained pairs from the final corpus. In each row the two branches face a
comparable local state; the positive and negative examples are the next actions
from the successful and localization branches, respectively. Thus, the mined
signal is not merely that the task is difficult: it identifies a local decision
for which the action changes and nearby trajectory evidence connects that
change to task progress or failure.

\begin{table*}[t]
\centering
\small
\setlength{\tabcolsep}{3pt}
\begin{tabular}{>{\raggedright\arraybackslash}p{0.20\textwidth}>{\raggedright\arraybackslash}p{0.16\textwidth}>{\raggedright\arraybackslash}p{0.19\textwidth}>{\raggedright\arraybackslash}p{0.19\textwidth}>{\raggedright\arraybackslash}p{0.16\textwidth}}
\csname toprule\endcsname
Task and aligned state & Failure mode ($S_{\mathrm{total}}$) & Successful-side action & Localization-side action & Why the step is critical \\
\midrule
Toyonaka care-insurance research; both branches are on a blocked search-results page & Target discovery or search-strategy failure (0.84) & Navigate directly to the official municipal website & Issue another query through the blocked search path & The direct pivot exits the blocker; repeating search leaves the run stalled. \\
Edit a submitted SEEK application; both branches show the same sign-in panel & Critical action or task endpoint incomplete (0.82) & Activate the sign-in control & Scroll away from the authentication panel & Authentication is the immediate prerequisite for reaching submitted applications. \\
Research selling a car through Carvana; both branches see the same Cloudflare block & Access, anti-bot, or technical obstruction (0.84) & Open an alternative search engine & Only write a planning file & One action resumes browser progress; the other changes no browser state. \\
Verify current US YouTube Premium prices; both branches see generic, region-mismatched results & Hard constraint, target object, or numeric threshold mismatch (0.76) & Rewrite the query to target official US support pages and submit it & Continue scrolling the generic result set & The query rewrite enforces both source and region constraints at the decision point. \\
\bottomrule
\end{tabular}
\caption{Real mined action contrasts from the retained corpus. The
successful-side action is the positive training target; the localization-side
action is shown only to audit why the selected state is failure-critical.}
\label{tab:critical-step-cases}
\end{table*}

{}\textbf{Optimization interface.} The contrast disappears after mining. For
each task, the miner retains only the selected successful-side observation and
its verified action as an RL prompt--target pair. GRPO samples candidate actions
from that observation and scores them against the verified successful action.
Neither the localization-side observation nor its action appears in the
policy-gradient objective; the localization branch remains in intermediate
artifacts only for reproducibility and auditing.


\fi
\ifdefined\SUPPLEMENTONLY
\bibliography{aaai2027}
\fi
\end{document}